%% file: main.tex
\documentclass[10pt,twocolumn,letterpaper]{article}

\usepackage{cvpr}              
\usepackage{sectsty}

\input{preamble}

\definecolor{cvprblue}{rgb}{0.21,0.49,0.74}
\usepackage[pagebackref,breaklinks,colorlinks,citecolor=cvprblue]{hyperref}
\usepackage{amsmath}
\def\paperID{3239} 
\def\confName{CVPR}
\def\confYear{2024}

\title{BusReF: Infrared-Visible images registration and fusion focus on reconstructible area using one set of features}

\author{Zeyang Zhang,Hui Li,Tianyang Xu,Xiaojun Wu\\
Jiangnan University\\
No.1800 Lihu Avenue, Wuxi City, Jiangsu Province, China\\
{\tt\small zzy\_jnu\_cv@163.com}
\and
Josef Kittler\\
University of Surry\\
Guildford, Surrey GU2 7XH, United Kingdom\\
{\tt\small j.kittler@surrey.ac.uk}
}

\begin{document}
\maketitle
\input{sec/0_abstract}    
\input{sec/1_intro}
\input{sec/2_Relatedwork}

\input{sec/3_Methods}
\input{sec/4_Experiments}

{
    \small
    \bibliographystyle{ieeenat_fullname}
    \bibliography{main}
}

\end{document}

%% file: preamble.tex
\usepackage[dvipsnames]{xcolor}


%% file: sec/0_abstract.tex
\begin{abstract}
In a scenario where multi-modal cameras are operating together, the problem of working with non-aligned images cannot be avoided. Yet, existing image fusion algorithms rely heavily on strictly registered input image pairs to produce more precise fusion results, as a way to improve the performance of downstream high-level vision tasks. In order to relax this assumption, one can attempt to register images first. However, the existing methods for registering multiple modalities have limitations, such as complex structures and reliance on significant semantic information. This paper aims to address the problem of image registration and fusion in a single framework, called BusRef. We focus on Infrared-Visible image registration and fusion task (IVRF). In this framework, the input unaligned image pairs will pass through three stages: Coarse registration, Fine registration and Fusion. It will be shown that the unified approach enables more robust IVRF. We also propose a novel training and evaluation strategy, involving the use of masks to reduce the influence of non-reconstructible regions on the loss functions, which greatly improves the accuracy and robustness of the fusion task. Last but not least, a gradient-aware fusion network is designed to preserve the complementary information.
The advanced performance of this algorithm is demonstrated by comparing it with different registration/fusion algorithms.

\end{abstract}

%% file: sec/1_intro.tex
\section{Introduction}
Image fusion is an important technique in the field of computer vision. The main purpose of image fusion is to generate fused images by integrating complementary information from multiple source images of the same scene, with the aim to improve the performance of downstream high-level semantic vision tasks~\cite{ZHANG2021323,9812535,8580578}. 

Typically, image fusion involves multiple sensors to capture scene data and the use of fusion algorithms to integrate the complementary information~\cite{9416507}.
The normal prerequisite for multi-modal image fusion is that the input image pairs are strictly aligned. The impact of misalignment is severe ghosting~\cite{Xu_2022_CVPR}. However, in most scenarios, due to the difference between  the internal and external parameters of infrared cameras and digital RGB cameras, 
it is very hard to obtain strictly aligned multi-modal image pairs. 
\begin{figure}
    \begin{center}
        \includegraphics[width=1\linewidth]{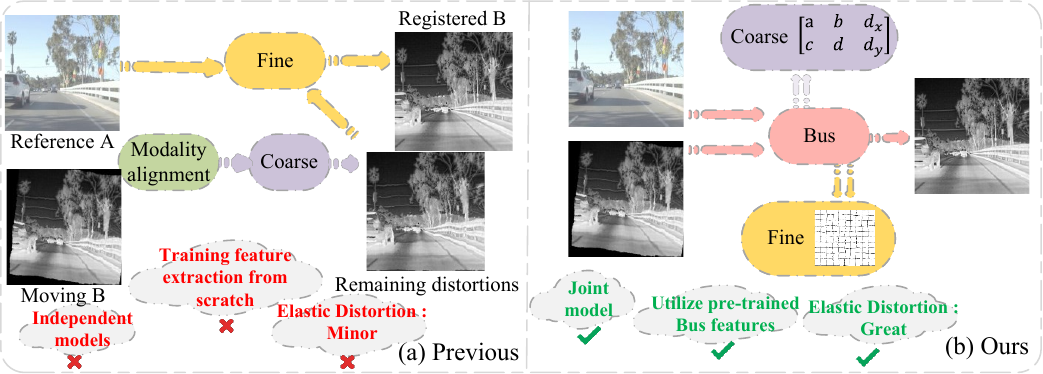}
        \caption{Comparison of frameworks. Existing serial training methods and our proposed Bus like training.}\label{overview}
    \end{center}
\end{figure}
To remedy this situation, it is crucial to perform the registration of multiple images before the fusion task. 

Due to the difference in imaging principles between infrared and digital cameras, feature-point matching-based registration algorithms 
often achieve unsatisfactory results. 
In most cases, the gradients of the salient target edges have a negative correlation between the two modalities, and sometimes the salient targets are even lost due to factors such as smoke occlusion. In these cases, most similarity operators lose effectiveness and even mislead the image registration.

In general, there are three existing solutions for Infrared - Visible image registration$:$
\begin{itemize}
    \item Specially designed similarity loss~\cite{Kong_2023_CVPR}
    \item Search for identical information between modalities~\cite{Xu_2022_CVPR,10145843,wang2022unsupervised}
    \item Semantic based methods~\cite{xie2023semantics}
\end{itemize}
Most of the existing algorithms contain multiple stages, with no interdependence between them. 
For example, a method may include a modality alignment module, a coarse registration module, and a fine registration module, which are trained~\cite{10145843,Xu_2022_CVPR,wang2022unsupervised,xie2023semantics} independently. However, this one-at-a-time training approach from scratch is very unsatisfactory, as it limits the feature extraction capabilities of each model. It ignores the fact that the image registration task often requires fine-grained structural descriptions~\cite{10145843}. 

In this paper, we design a unified framework for multi-modal image fusion, linking coarse and fine registration modules through a backbone. The unique features of our framework are shown in the comparison with conventional approaches in Figure \ref{overview}. The representation extracted by the backbone network provides the fine-grain features to sub-modules for registration. Specifically, an Auto-Encoder (AE) framework~\cite{9127964,LI202172} is trained as the bus of our proposed algorithm. Once the training is completed, the parameters of AE are fixed, and the design proceeds with training the coarse registration module~\cite{LI202326}, fine registration module~\cite{Arar_2020_CVPR}, and multi-modal image fusion module~\cite{CHENG202380,9550760,9018389,10105495} mounted on the bus (BusReF).

In our approach, we also address the problem of fusing unaligned multi-modal images subject to content inconsistencies. In many existing algorithms, one assumes that the information from one modality is used as a reference and the other as a moving image~\cite{Kong_2023_CVPR}. Under this assumption, the registration methods tend to resample the moving image on a grid to match the reference image. 
However, due to the difference in the perspective projections of multimodal cameras, resulting in incongruent magnifications, some of the information in the reference images, compared to the moving images, during the un-warping process may be missing. It is worth noting that most of the "missing information" continuously affects the reference image, constituting "non-reconstructible" regions. Existing algorithms often ignore the impact of these "non-reconstructible" regions on the registration problem, and their model will forcibly attempt to register these parts regardless. 
To address this problem, we propose a reconstructible mask and apply it to the loss function during the training process, which greatly improves the registration ability of the model and reduces the risk of false matches. 

Finally, in order to improve the performance further, the task of image fusion and registration should be in a mutually reinforcing relationship~\cite{10145843}. We propose a gradient-aware fusion network(GAF) and use it as guidance during the training phase of image registration.

%% file: sec/2_Relatedwork.tex
\section{Related Work}
Multi-modal image registration has been widely discussed mainly in the field of medical images~\cite{hill2001medical,ZHOU2023134,DINH2023104740}. Due to the differences in imaging principles between modalities, the appearance of a target rendered by different modalities may vary greatly. In general, solving the image registration problem involves two major approaches: i) finding similar features between modalities and computing a spatial transformation based on them, ii) learning a spatial transformation to maximise the similarity between the modalities.
\subsection{Multi-Modality Registration}
Nowadays there are many methods based on feature matching. SIFT~\cite{lowe2004distinctive} is used as a local feature description operator. 
LoFTR~\cite{Sun_2021_CVPR} innovatively combines CNN and Transformer for feature matching~\cite{NEURIPS2018_8f7d807e,10.1007/978-3-030-58545-7_35,NEURIPS2020_c91591a8,dosovitskiy2020image}. 
MatchFormer~\cite{Wang_2022_ACCV} leverages the powerful global feature interaction capabilities of the Transformer to perform multi-scale feature matching through a hierarchical structure. 
However, the aforementioned algorithms are designed for single modality image pairs and the performance will be seriously degraded when directly applied to processing multi-modal unaligned image pairs. Due to the different imaging principles of different sensors, the corresponding regions of different images may exhibit only a weak correlation. When it comes to Thermal Infrared–Visible (TIR) image pairs, the gradients of salient target regions at the edges are likely to show a negative correlation. Sometimes the target will be completely invisible. Semantics led all (SemLA)~\cite{xie2023semantics} achieves robust feature matching in simple scenes by using the results of semantic segmentation and restricting the feature matching to significant semantic target regions.

Many works seek a representation that is shared by multiple modalities. Cross-modality Perceptual Style Transfer Network (CPSTN)~\cite{wang2022unsupervised} was the first method to apply CycleGAN~\cite{Zhu_2017_ICCV} to perform style transformation and reduce inter-modality differences.
Mutually Reinforcing Multi-modal Image Registration and Fusion (MURF)~\cite{10145843,NEURIPS2020_d6428eec} has a similar idea. The only difference is that it uses contrastive learning to learn a robust and unified description of multi-modality images.

The quest for a unified description of multi-modal data is challenging. In most cases the intermediate features are less detailed, making it difficult to achieve high accuracy in learning multi-modal registration. Another promising approach has been suggested by Li et al.~\cite{LI202326} who directly and dynamically align the features of the convolutional network and depends on these features for image reconstruction. 

However, all existing methods suffer from low coupling between the modules, task fragmentation and poor feature reusability. Finding intermediate features, performing a coarse registration, refining it and finally conducting image fusion are all formulated as unrelated multiple tasks, with multiple models to solve (rather than sub-modules within a model), and trained separately. This serial training approach makes it necessary for each of the models to learn feature descriptions from scratch, and it is difficult to ensure that the features learned by each model are optimal for different scenarios.

\subsection{Multi-Modality Fusion}

Multi-modal image fusion requires a high-precision alignment of image pairs. Otherwise, artefacts will appear in the fused image, hindering subsequent high-level semantic tasks~\cite{Zhao_2023_CVPR}. The fusion algorithm should be able to preserve complementary information between multi-modal images and present it clearly in the fused image. For infrared-visible fusion, the widely accepted definition of complementary information is that the fused image should reflect the texture details of the visible image and the highlighted salient targets in the infrared image~\cite{zhao2020didfuse}.

In deep learning-based image fusion algorithms, DenseFuse~\cite{8580578} pioneered the application of Auto-Encoder, while encouraging feature reuse and specially designed feature fusion rules to improve the performance of image fusion dramatically. In recent years, end-to-end fusion also has gained wide attention due to its limited dependence on human heuristics, and its high robustness.

In this paper, a multi-modal image registration and fusion network that imitates a bus structure is proposed. Specifically, our multi-modality registration network has an Auto-Encoder as a bus, and the sub-modules required for registration are mounted on the bus one by one for cooperative training. Since the Auto-Encoder requires features that can reconstruct the original image, they are inevitably rich in texture details, and therefore well suited for the registration task. Other innovations of our approach are the proposed reconstructible region mask, a gradient-aware fusion network, and a fusion strategy for registration training, designed to provide more robust registration and better visual results for unaligned multi-modal image fusion.

%% file: sec/3_Methods.tex
\section{Methods}
\subsection{Reconstructible Mask}

\begin{figure}
    \begin{center}
        \includegraphics[width=1\linewidth]{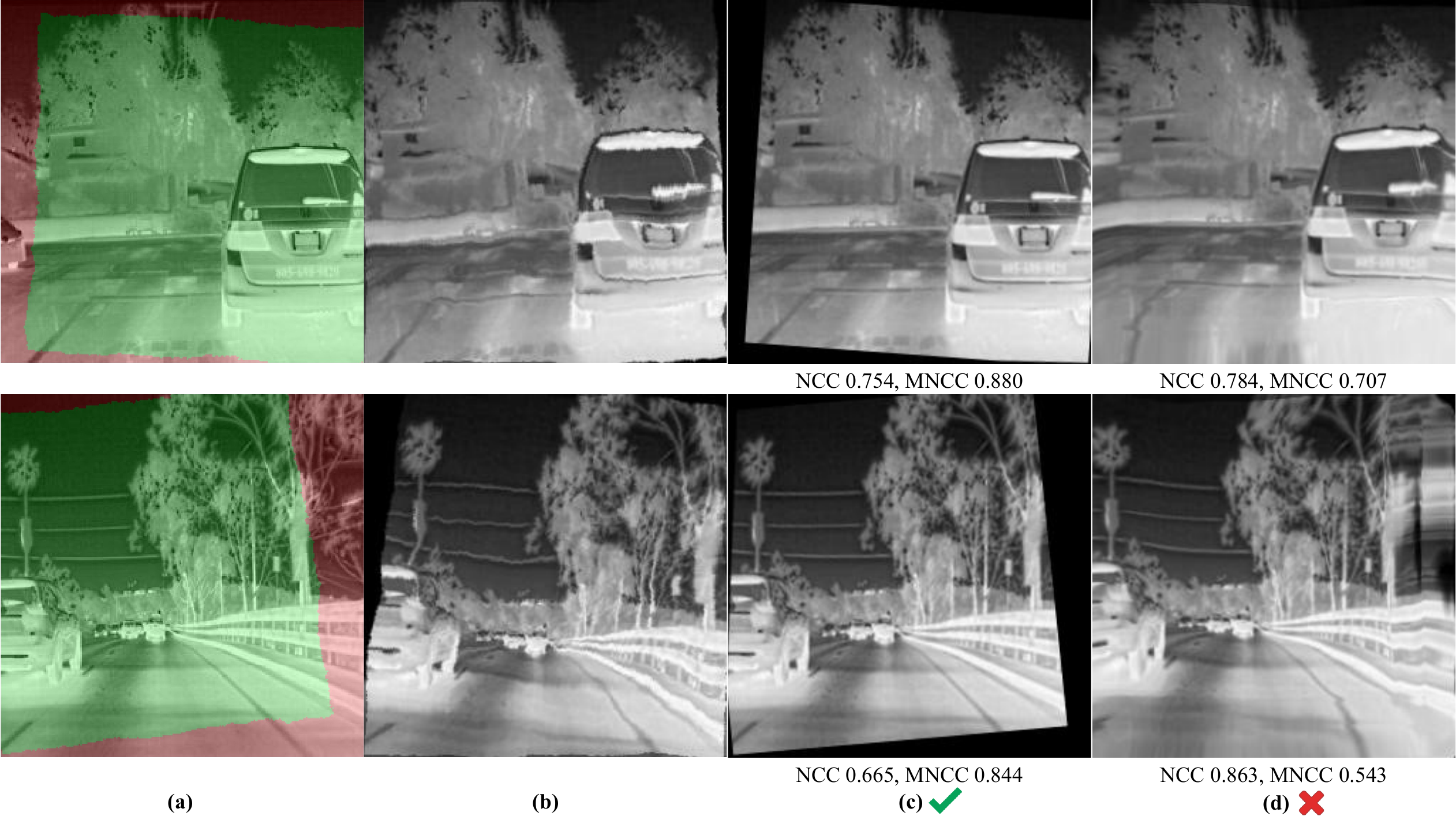}
        \caption{(a) Original image. The red mask represents the unreconstructible area, the green represents our reconstructible content mask. (b) Artificially generated unaligned image. (c) Ground truth registration results. (d) Possible artifacts of not applying the reconstructible mask.}\label{gradmask}
    \end{center}
\end{figure}

Due to the differences in intrinsic and extrinsic parameters of different sensors, it is normal that captured image pairs are not fully registered. As shown in Figure~\ref{gradmask}, the red portion represents the area of missing information in the moving image.

In this work, we propose the use of a reconstructible mask, which is used to guide the network to focus on image information that can be registered. Let us assume that $I_x$ and $I_y$ are a strictly registered multi-modal image pair. $\theta$ is a randomly generated 6-dof affine transformation parameters~\cite{NIPS2015_33ceb07b} to simulate the rigid deformation between the images, while the deformation field $\phi$ is to simulate the elastic deformation~\cite{hellier2001hierarchical}. See the appendix for details of the generation process.
\begin{equation}
    \theta = \begin{bmatrix}
        a & b & d_x \\
        c & d & d_y
    \end{bmatrix},
    \theta^{-1} = \begin{bmatrix}
        a & b \\
        c & d
    \end{bmatrix}^{-1}
    \begin{bmatrix}
        -d_x \\
        -d_y
    \end{bmatrix}
\end{equation}
\begin{equation}
    \phi^{-1} = -\phi
\end{equation}

Then the spatial mapping between the images can be defined by an affine transformation $S()$. The spatially transformed image $I_y$ is $I^R_y = S(I_y,\theta)$. The deformation field can be modelled by elastic deformation, denoted as $E()$, the elastic transformed $I^R_y$ is $I^{RF}_y = E(I^R_y,\phi)$. Here $S()$ and $E()$ are grid sample functions. By this formulation, we have obtained an artificially unaligned image pair and forward and backward transforms. For strictly aligned images, the reconstructible region would be the full image. Hence, the mask $M$ can be initialised as a matrix of ones, of size  $I_x, I_y$. The forward transformation is applied to $M$:

\begin{equation}
\begin{aligned}
     &M^{RF} = E(S(M,\phi),\theta),\\
     &M^{\prime} = E(S(M^{RF},\phi^{-1}),\theta^{-1}),\\
     &\Bar{M} = \begin{cases}
     1, &   M^{\prime}>0, \\
     0, &   else.
     \end{cases}
\end{aligned}
\end{equation}
 where $\Bar{M}$ represents the reconstructible area between the reference images and moving images.
 
\subsection{Bus Like Training Strategy}
\begin{figure*}
    \begin{center}
        \includegraphics[width=0.9\linewidth]{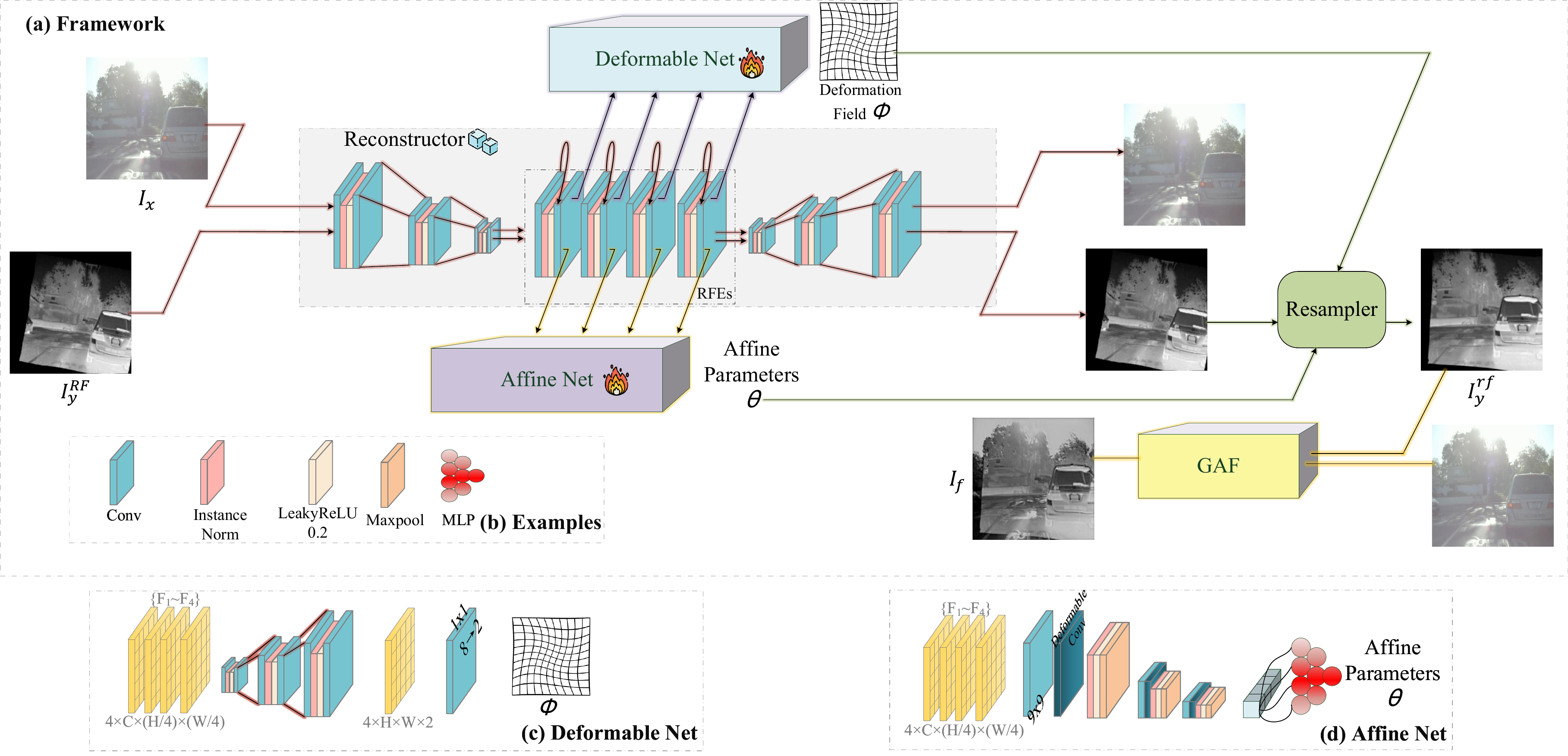}
        \caption{The Reconstructor is an Auto-Encoder framework that is trained by simultaneously inputting Infrared-Visible images and requiring the reconstruction of the input images to ensure the ability to extract multi-modal features. The registration module is mounted on this pre-trained framework to ensure the acquisition of detailed features. Finally, the affine transformation parameter $\theta$ and the deformation field $\phi$ corresponding to the rigid and elastic transforms are learnt. Finally, the mesh resampling is performed to achieve the image registration.}\label{pipeline}
    \end{center}
\end{figure*}
Multi-modality image registration requires the inputs to contain detailed information, which is difficult to achieve by training each module from scratch. As shown in Figure ~\ref{pipeline}, Auto-Encoder is used as the bus in our framework, the features are extracted by Encoder from source images which are also reconstructed by Decoder with these features. This processing can ensure that all the detailed information is retained in the extracted feature maps. The Auto-Encoder as a reconstructor contains three down-sampling modules, four residual feature extraction modules (RFE) and three up-sampling modules. Four RFEs are denoted as $\{R_1\sim R_4\}$, and their outputs are $\{F_1\sim F_4\}$, representing the features from shallow to deep, respectively. 
The loss function for the reconstructor is composed of the $l_2$ norm based pixel loss and structural similarity index measure(SSIM) loss~\cite{1284395},
\begin{equation}
    \mathcal{L}_{re} = \| I - I_{re}\|_2 + \lambda (1-SSIM(I,I_{re}))
\end{equation}
where $\lambda$ is a balancing hyperparameter, making the two loss functions to be of similar magnitude, $I_{re}$ is the reconstructed image.

To make the registration server better for multi-modal image fusion, the fusion module should be trained in coordination with the registration modules to enhance the performance of each module. When edges in the multi-modal images perfectly overlap in the fused image, it represents a highly accurate alignment.

\begin{figure*}
    \begin{center}
        \includegraphics[width=0.85\linewidth]{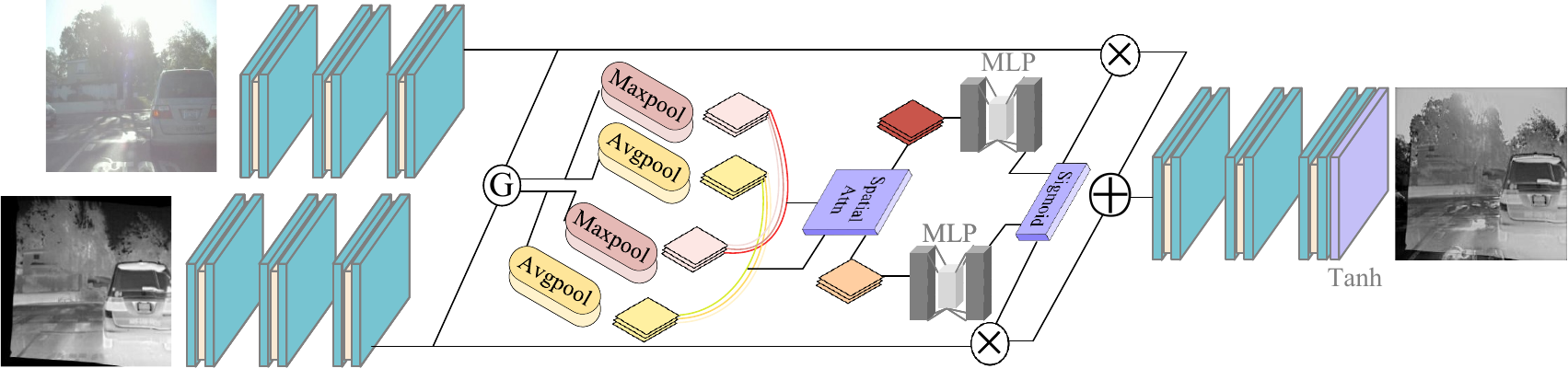}
        \caption{The architecture of GAF. The registered image pairs are inputted to the feature extractor, and gradient sensing is performed on the extracted features. $G$ is the Laplacian operator. After separating the high-frequency part and low-frequency part by Maxpool and Avgpool respectively, 
        two MLPs are used to learn inter-modality weighting.
        Finally, the fused image is obtained by output   convolutional layers and Tanh activation.}\label{fusion_network}
    \end{center}
\end{figure*}

Gradient information tends to reflect edges in the images.  The greater the consistency between the edges, 
 the better the registration. Therefore the fusion network should be able to gauge the gradient. As shown in Figure ~\ref{fusion_network}, the spatial attention is utilized to establish the gradient map, based on the existing gradient-weighted fusion network~\cite{10145843}. The loss function of the fusion network includes a weighted SSIM $\mathcal{L}_{wsim}$ and gradient loss $\mathcal{L}_{grad}$.
\begin{equation}
    \mathcal{L}_{wsim} = 1-\frac{{SSIM}(I_f,I_v)+{SSIM}(I_f,I_r)}{2}
\end{equation}
where $I_f$ is the fused image, $I_v$ is the visible light image and $I_r$ is the infrared light image. 

$\mathcal{L}_{grad}$ requires $I_f$ and $I_t$ to have similar gradient information:
\begin{equation}
    \mathcal{L}_{grad} = \|\nabla I_f-\nabla I_t \|_2
\end{equation}
where $\nabla$ is the laplacian operator. The target gradient we want to retain is $\nabla I_t$:

    \begin{align}
            &\nabla I_t = w(\frac{\nabla I_v\cdot{\vert\nabla I_v\vert}^\gamma}{\vert\nabla I_v\vert})+(1-w)(\frac{\nabla I_r\cdot{\vert\nabla I_r\vert}^\gamma}{\vert\nabla I_r\vert})\nonumber\\
            s.t.\quad &w[i,j] = \begin{cases}
        1, &   \vert\nabla I_x[i,j]\vert > \vert\nabla I_y[i,j] \vert \\
        0, &   else.
     \end{cases}
    \end{align}

where $\gamma$ is the enhancement factor, which is set to 0.7.

The loss function of the fusion network is the weighted summation of $\mathcal{L}_{wsim}$ and $\mathcal{L}_{grad}$
\begin{equation}
    \mathcal{L}_{fuse} = \sigma\mathcal{L}_{wsim} + \mathcal{L}_{grad}
\end{equation}

Once the reconstruction and fusion network is designed, we treat it as a bus for the framework and freeze the parameters in it. Then we mount Affine Net for learning coarse registration. The input unaligned image pairs are first processed $\{F_1\sim F_4\}$ by the reconstructor and used as the input to Affine Net. Multiple down-sampling processes are included in Affine Net and before each down-sampling we use a large scale $7\times7$ convolution and a dynamic convolution to achieve a sufficiently large receptive field~\cite{Chen_2020_CVPR}. Finally, the global pooling and MLP are used to obtain the desired affine transformation parameter $\theta$.

At the same time, we mount up Deformable Net for elastic deformation learning. The input of Deformable Net is $\{F_1\sim F_4\}$. It is worth noting that the output layer of Deformable Net is a simple $3\times3$ convolution with 8 input channels and 2 output channels. We perform group convolution~\cite{Zhang_2017_ICCV} on $\{F_1\sim F_4\}$ to ensure that each $F_i$ only outputs 2 channels. This is equivalent to each layer of the RFE giving an elastic deformation proposal. Finally, we use a convolutional layer to linearly weight these 4 deformation proposals to output the final Deformation Field.

\begin{equation}
    \begin{aligned}
    &I_y^{Rf} = E(I_y^{RF}, A(\{F_1\sim F_4\}_x,\{F_1\sim F_4\}_y)),\\
    &I_y^{rf} = S(I_y^{Rf}, D(\{F_1\sim F_4\}_x,\{F_1\sim F_4\}_y))
    \end{aligned}
\end{equation}
where $A()$ is Affine Net, $D()$ is Deformable Net and $I_y^{rf}$ is the reversed transformation result. During the training of the registration module, $A()$ and $D()$ are simultaneously mounted on the bus, and the loss function at this stage is constituted by the masked NCC loss $L_{MNCC}$ and the masked gradient loss $L_{MG}$~\cite{1660446}.

\begin{equation}
    \mathcal{L}_{MNCC} = -MNCC(I_y^{rf},I_y,\Bar{M})
\end{equation}

\begin{equation}
\small
    \begin{aligned}
        &MNCC(x,y,\Bar{M})=\\
        &\frac{\sum\limits_{i}^{i\in\Bar{M}=1}\sum\limits_{j}^{j\in\Bar{M}=1} (x_{i,j}-\Bar{x})(y_{i,j}-\Bar{y})}
        {\sqrt{\sum\limits_{i}^{i\in\Bar{M}=1}\sum\limits_{j}^{j\in\Bar{M}=1}(x_{i,j}-\Bar{x})^2}\sqrt{\sum\limits_{i}^{i\in\Bar{M}=1}\sum\limits_{j}^{j\in\Bar{M}=1}(y_{i,j}-\Bar{y})^2}}
    \end{aligned}        
\end{equation}
where $x,y$ indicate two images. $\Bar{x},\Bar{y}$ denote their mean values.
\begin{equation}
    \mathcal{L}_{MG} = \|(\nabla F(I_x,I_y^{rf})- \nabla F(I_x,I_y))\cdot\Bar{M}\|_2
\end{equation}
where $F()$ is the GAF. The loss function to train registration is $\mathcal{L}_{reg}$
\begin{equation}
    \mathcal{L}_{reg} = \epsilon\mathcal{L}_{MNCC} + \mathcal{L}_{MG}
\end{equation}

%% file: sec/4_Experiments.tex
\section{Experiments}
Unaligned multi-modality image registration and fusion is a very challenging task and there are a few competitive works to compare with BusReF. In this section, some single modality registration algorithms such as LoFTR and SIFT operator are chosen for comparison, using GAF to perform the fusion. The algorithms compared include also some representative state-of-the-art multi-modality registration and fusion algorithms such as SemLA and MURF.

\subsection{Qualitative Comparison}
\begin{figure*}
    \begin{center}
        \includegraphics[width=1\linewidth]{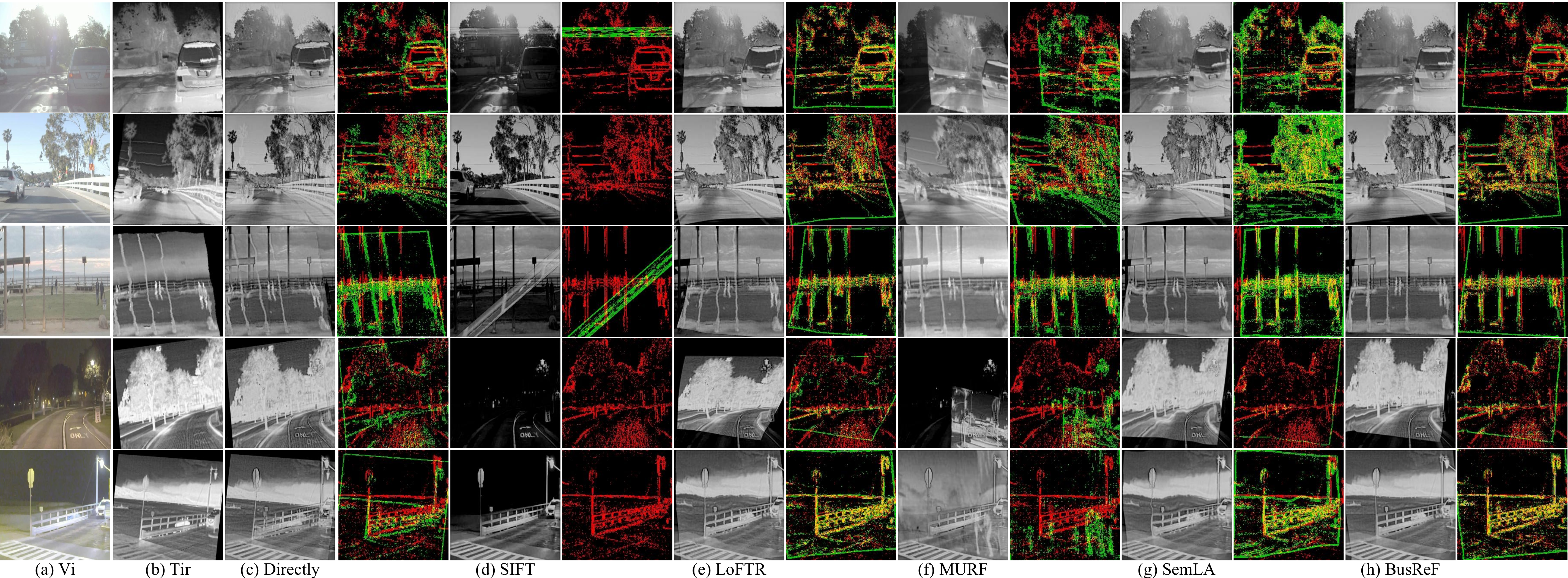}
        \caption{A qualitative comparison of fusion and registration on the TIR task (RoadScene) dataset. (a), (b) the original visible image and thermal Infrared image, respectively. (c) the directly fused results obtained by GAF. (d) SIFT+GAF. (e) LoFTR+GAF. (f), (g) the results obtained by MURF and SemLA. (h) the results of BusReF.}
        \label{compare_register_tir}
    \end{center}
\end{figure*}

\begin{figure*}
    \begin{center}
        \includegraphics[width=1\linewidth]{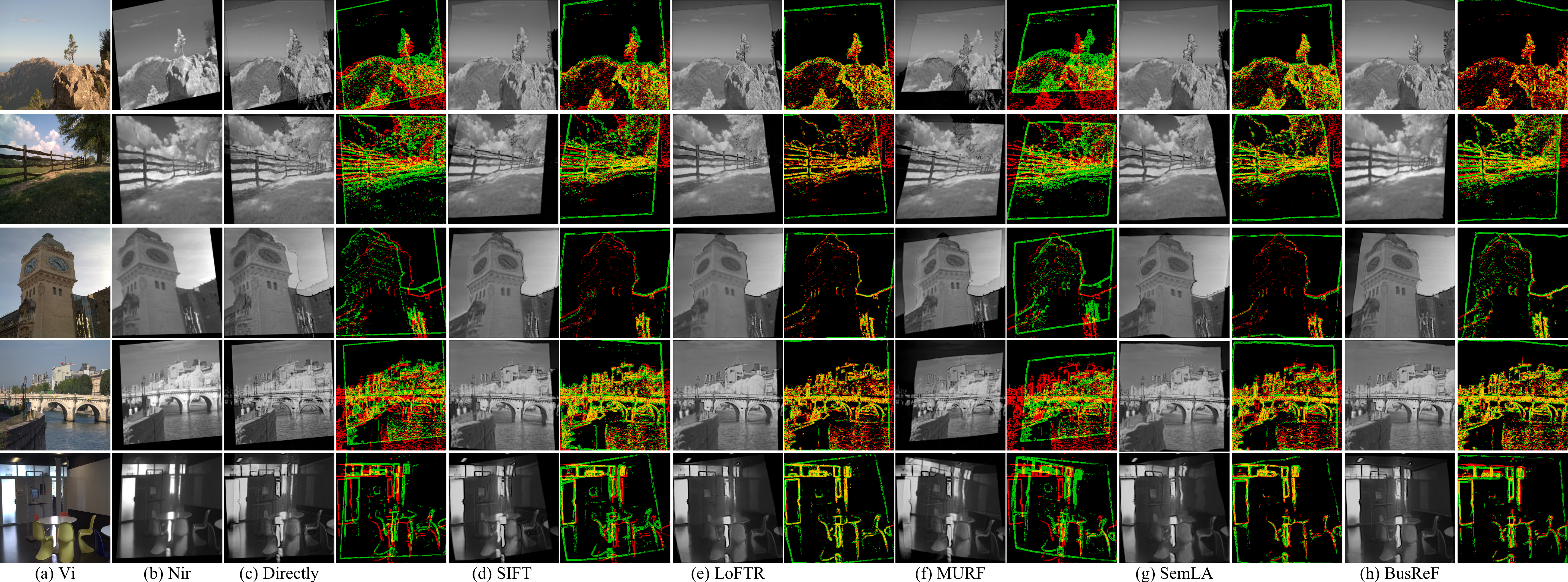}
        \caption{A qualitative comparison of fusion and registration on the NIR task. The setup of the experiment was kept consistent with Figure~\ref{compare_register_tir}.}\label{compare_register_nir}
    \end{center}
\end{figure*}

\begin{figure*}
    \begin{center}
        \includegraphics[width=1\linewidth]{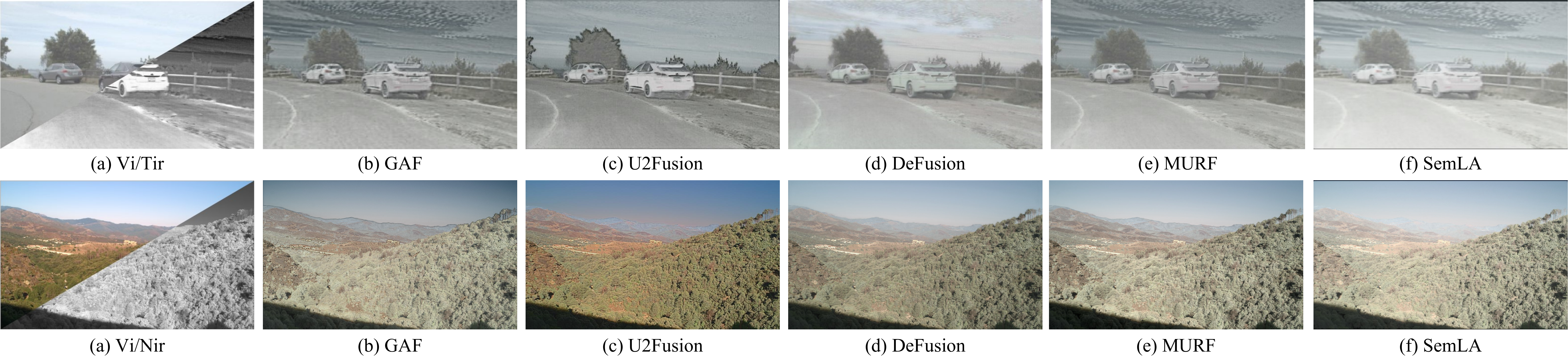}
        \caption{A qualitative comparison of the GAF with state-of-the-art multi-modality fusion algorithms. The left set is from the RoadScene dataset, showing a car parked by a railing, and the right set is from the NIRScene dataset showing a mountain scene. (a) the original images from the datasets (b) the GAF fusion result. (c),(d) the output of state-of-the-art unified image fusion algorithms. The (e),(f) The two works include both registration and fusion modules, we only show the results of their fusion modules here.}\label{qualitative_fusion}
    \end{center}
\end{figure*}

The TIR registration and fusion results are given in Figure~\ref{compare_register_tir}. The red and green edge salient maps are the superimposed gradient maps of the registered IR images and the human manually registered images. 

As seen in Figure~\ref{compare_register_tir} (c) and its edge salient maps, a direct fusion of unaligned multi-modality images produces many artefacts, resulting in the injection of additional noise instead. The SIFT operator and LoFTR are designed for unimodal unaligned images, where SIFT is completely incapable of solving the TIR-VI task, whereas LoFTR, thanks to the deep features extracted by Convolution and Transformer networks, is able to perform the registration in some of the cases. 

Observing the results of the fourth row in Figure~\ref{compare_register_tir}, the woods in the image are highlighted in the infrared modality and almost completely black and invisible in the visible modality, which shows a negative correlation trend. In contrast, the traffic lines and symbols on the road belong to the same highlighted information, and their gradients show a positive correlation. 

It can be seen that LoFTR based on feature matching is able to register positively correlated information but not negatively correlated information. SemLA belongs to the family of algorithms based on feature matching, but some regions of the registered images are too distorted. The proposed method, BusReF, predicts the affine transformations in the TIR-VI task more accurately and can correct most of the elastic deformations. Moreover, BusReF is also able to perform high-precision registration in dark scenes, such as the penultimate and penultimate rows of Figure~\ref{compare_register_tir}.

Figure~\ref{compare_register_nir} shows the registration and fusion results of NIR-VI. Five representative scenes are selected for a qualitative comparison. Starting from the first row of Figure~\ref{compare_register_nir} shows "Mountain", "Country", "Old-Building", "Water", "Indoor" scenes. The differences between NIR-VI modalities are small, and there are fewer areas of negative correlation between the NIR-VI image pairs. Therefore most of the algorithms are able to complete the alignment to some extent. However, as seen by visualising the gradient salient maps, BusReF results in fewer artefacts and better edge overlap. 

When evaluating the fusion performance of GAF for the thermal infrared modality, as can be seen from the image of the car parked by the guardrail in Figure~\ref{qualitative_fusion}, the fusion result of GAF has sharper edges and the clouds in the sky are well preserved. The thermal infrared information of the TIR modality is also well represented in the fused image.

We also give the NIR-VI fusion results of the mountains in Figure~\ref{qualitative_fusion}. Among all the compared algorithms, the GAF's fusion image has the clearest distant mountains and the infrared radiation information of the near mountains is well represented. The result obtained by U2Fusion produces a better view, but the distant mountains become transparent and invisible after fusion. The SemLA fusion was the brightest, but too close to the TIR image and the distant mountains were also less visible.

\subsection{Quantitative Comparison}
\begin{table}[]
\centering
\caption{A comparison of the registrations of 221 image pairs on the RoadScene dataset and 469 image pairs on the NIRScene dataset. {\color[HTML]{FE0000}{\textbf{Red}}} bold is the best, {\color[HTML]{3166FF}{\textbf{blue}}} bold is the second.}\label{registration}
\label{compare_reg}
\resizebox{\columnwidth}{!}{%
\begin{tabular}{c|ccccc}
\hline
\begin{tabular}[c]{@{}c@{}}RoadScene\\ (TIR)\end{tabular} & SIFT & LoFTR & \begin{tabular}[c]{@{}c@{}}SemLA\\ (Matchformer)\end{tabular} & MURF & BusReF \\ \hline\hline
NCC & 0.113±0.277 & 0.801±0.034 & \color[HTML]{3166FF}{\textbf{0.820±0.023}} & 0.442±0.111 & \color[HTML]{FE0000}{\textbf{0.876±0.028}} \\
MSE$\downarrow$ & 0.256±0.009 & 0.059±0.001 &  \color[HTML]{FE0000}{\textbf{0.040±0.004}} & 0.106±0.010 & 
\color[HTML]{3166FF}{\textbf{0.042±0.005}} \\
MNCC & 0.073±0.024 & 0.755±0.019 & \color[HTML]{3166FF}{\textbf{0.813±0.007}} & 0.373±0.071 &  \color[HTML]{FE0000}{\textbf{0.916±0.002}} \\
MMSE$\downarrow$ & 0.228±0.007 & 0.035±0.006 & \color[HTML]{3166FF}{\textbf{0.025±0.000}} & 0.110±0.007 &  \color[HTML]{FE0000}{\textbf{0.011±0.005}} \\ \hline
\begin{tabular}[c]{@{}c@{}}NIRScene\\ (NIR)\end{tabular} & SIFT & LoFTR & \begin{tabular}[c]{@{}c@{}}SemLA\\ (Matchformer)\end{tabular} & MURF & BusReF\\ \hline\hline
NCC &  {0.345±0.117} &  {0.832±0.036} &  \color[HTML]{3166FF}{\textbf{0.851±0.033}} &  {0.796±0.042} &   \color[HTML]{FE0000}{\textbf{0.869±0.040}} \\
MSE$\downarrow$ &  {0.125±0.018} &  \color[HTML]{3166FF}{\textbf{0.029±0.000}} &   \color[HTML]{FE0000}{\textbf{0.028±0.000}} &  {0.097±0.002} &  {0.030±0.000} \\
MNCC &  {0.361±0.098} &  \color[HTML]{3166FF}{\textbf{0.647±0.034}} &  {0.619±0.032} &  {0.450±0.027} &   \color[HTML]{FE0000}{\textbf{0.897±0.005}} \\
MMSE$\downarrow$ &  {0.117±0.142} &  {0.353±0.006} &  \color[HTML]{3166FF}{\textbf{0.036±0.001}} &  {0.078±0.001} &   \color[HTML]{FE0000}{\textbf{0.009±0.004}} \\ \hline
\end{tabular}%
}
\end{table}
\begin{table}[]
\centering
\caption{A comparison of the fusion of 221 image pairs on the RoadScene dataset and 469 image pairs on the NIRScene dataset.}\label{fusion}
\label{compare_fusion}
\resizebox{\columnwidth}{!}{%
\begin{tabular}{c|ccccc}
\hline
\begin{tabular}[c]{@{}c@{}}RoadScene\\ (TIR)\end{tabular} & DeFusion & SemLA & U2Fusion & MURF & GAF \\ \hline\hline
EI & 35.06±5.7 & 48.98±7.6 & 50.49±10.9 & {\color[HTML]{3166FF} \textbf{50.59±11.2}} & {\color[HTML]{FE0000} \textbf{63.98±7.0}} \\
CE$\downarrow$ & 0.91±0.1 & 0.93±0.1 & 0.99±0.2 & {\color[HTML]{3166FF} \textbf{0.81±0.3}} & {\color[HTML]{FE0000} \textbf{0.79±0.2}} \\
SF & 9.16±1.1 & {\color[HTML]{FE0000} \textbf{18.76±2.5}} & 12.37±1.8 & 15.36±3.6 & {\color[HTML]{3166FF} \textbf{17.45±2.4}} \\
$FMI_w$ & 0.27±0.0 & {\color[HTML]{3166FF} \textbf{0.36±0.1}} & {\color[HTML]{FE0000} \textbf{0.37±0.0}} & 0.26±0.0 & 0.24±0.0 \\
$Q_{cv}$ & 172.31±64.1 & {\color[HTML]{FE0000} \textbf{687.53±53.2}} & 255.47±25.5 & 497.67±43.6 & {\color[HTML]{3166FF} \textbf{556.34±35.8}} \\ \hline
\begin{tabular}[c]{@{}c@{}}NIRScene\\ (NIR)\end{tabular} & DeFusion & SemLA & U2Fusion & MURF & GAF \\ \hline\hline
EI & 57.97±12.3 & {\color[HTML]{3166FF} \textbf{69.04±14.5}} & 64.77±8.2 & 67.37±15.6 & {\color[HTML]{FE0000} \textbf{70.01±16.6}} \\
CE$\downarrow$ & {\color[HTML]{3166FF} \textbf{0.75±0.1}} & 0.98±0.2 & 0.90±0.1 & 0.88±0.3 & {\color[HTML]{FE0000} \textbf{0.72±0.3}} \\
SF & 18.05±5.7 & {\color[HTML]{FE0000} \textbf{25.02±6.8}} & 21.40±8.5 & 20.43±6.2 & {\color[HTML]{3166FF} \textbf{22.50±4.6}} \\
$FMI_w$ & {\color[HTML]{FE0000} \textbf{0.54±0.2}} & 0.41±0.1 & 0.34±0.0 & 0.37±0.0 & {\color[HTML]{3166FF} \textbf{0.48±0.1}} \\
$Q_{cv}$ & 659.47±160.3 & {\color[HTML]{333333} 845.23±210.5} & {\color[HTML]{FE0000} \textbf{1192.17±126.4}} & {\color[HTML]{3166FF} \textbf{946.89±266.2}} & 832.02±136.7 \\ \hline
\end{tabular}%
}
\end{table}

For objective evaluation of the BusReF registration capabilities, we give the results of quantitative experiments in Table~\ref{registration}. 

The first two rows of these two tables show the metrics computed on a full-map scale, with the higher NCC representing a higher correlation between the registration results and those manual registration results in the dataset. BusReF is marginally better in NCC and MSE metrics (around 0.01) than SOTA. 

Furthermore, observing the third and fourth rows of the two tables, MNCC and MMSE demonstrate that the calculation of these two metrics occurs only within the theoretically reconstructible area. In the TIR-VI task, the MNCC of BusReF is 0.1 higher than the SOTA with less variance, and BusReF leads in the NIR-VI task. Meanwhile, from Figure~\ref{gradmask}(c) and (d), a higher full-map NCC does not necessarily represent higher registration accuracy, and accordingly, the improvement of MNCC is more in line with human viewing.

The current SOTA fusion algorithms do not have multi-modality registration algorithms that can be used directly. In order to compare the performance of the fusion module GAF and the SOTA fusion algorithms, we input the strictly aligned image pairs from the original dataset into GAF and other fusion algorithms. The results are presented in Table~\ref{fusion}.

To evaluate the performance of multi-modal image fusion, we targeted EI~\cite{LUO201746}, SF~\cite{477498}, $Q_{cv}$~\cite{CHEN2007193}, $FMI_w$~\cite{7036000} and CE~\cite{shreyamsha2013multifocus} as evaluation metrics.  EI and SF  reflect the density and sharpness of the edges in the fused image, and $Q_{cv}$ reflects the image quality of the local region, calculated as the mean square error of the weighted difference between the fused image and the source image. $FMI_w$ denotes the degree of correlation between the two images, and CE analyses the degree of information retention between the fused image and the source image from an information theoretic perspective.

From Table~\ref{fusion}, GAF achieves two best values and two second-best values in  TIR-VI image fusion. The EI and SF show that the images fused by our algorithm have richer and sharper edges.  CE achieving the first place represents that the fusion result of GAF is able to retain more information of the source image and no additional noise is introduced. However, the $FMI_w$ metric appears not to be good in the fusion task of TIR-VI, but this is not supported by the results of the visualization. This is due to the fact that the fused image is able to reflect the information of both modalities at the same time. In the TIR-VI task, the value decreases when correlation calculations are performed with a single modality. 

Referring to NIR-VI, GAF achieves second place in the $FMI_w$ metric, which is not consistent with TIR-VI. This inconsistency arises from the magnitude of variability in the multi-modality data, with less variability between NIR-VI modes and more variability between TIR-VI modes. Thus the correlation between the fused images of the TIR-VI task and any single modality is small, whereas the correlation between the fused images of the NIR-VI task is considerable. This discrepancy reflects the ability of the GAF to embody multi-modal information simultaneously. The remaining metrics achieved two firsts and one second in the NIR-VI task, and the $Q_{cv}$ metric exhibited a small gap compared to the remaining algorithms.
\subsection{Ablation Study}
\begin{table}[]
\centering
\caption{A quantitative comparison of full BusReF without the reconstructible region masks, no Bus training, and lack of fusion task guidance. The data in the table are averaged over the TIR and NIR datasets for the registration results.}
\label{ablation}
\footnotesize
\begin{tabular}{c|clccc}
\hline
\begin{tabular}[c]{@{}c@{}}Avg\\ TIR \& NIR\end{tabular} & \multicolumn{2}{c}{Full} & Mask & \begin{tabular}[c]{@{}c@{}}Bus\\ Training\end{tabular} & \begin{tabular}[c]{@{}c@{}}Fusion\\ Guide\end{tabular} \\ \hline\hline
NCC & \multicolumn{2}{c}{\textbf{0.872}} & 0.852 & 0.574 & 0.829 \\
MSE {$\downarrow$} & \multicolumn{2}{c}{0.036} & \textbf{0.014} & 0.124 & 0.071 \\
MNCC & \multicolumn{2}{c}{\textbf{0.906}} & 0.638 & 0.418 & 0.839 \\
MMSE {$\downarrow$} & \multicolumn{2}{c}{\textbf{0.010}} & 0.087 & 0.201 & 0.036 \\ \hline
\end{tabular}%
\end{table}

We conducted ablation experiments investigating the three proposed improvement measures. In order to capture the impact of the method on the combined TIR and NIR tasks, the metrics we adopted were averaged over the two datasets. 

First, we use the same training strategy, but without using the Reconstructible mask in the loss function for training. As seen in the third column of Table~\ref{ablation}, NCC and MSE are almost indistinguishable from the full method, and the MSE metric is even better than the full method, but the MNCC and MMSE metrics decrease substantially. This means that networks trained without Reconstructible masks are likely to forcefully register the regions that cannot be reconstructed to optimise the loss functions. This can lead to regions that should be interested not being handled well. 

In the second ablation study, the registration modules Affine Net and Deformable Net were removed from the bus for serial individual training. Specifically, the unaligned multi-modal image pairs are fed to the registration modules, which output the transformation parameters to register infrared images. As seen in the fourth column of Table~\ref{ablation}, all four metrics show a substantial improvement, but the network is unable to complete registration. This is because the registration modules require a lot of capacity for learning an adaptive feature representation. This implies that the difficult task of multi-modality registration can not be learned. 

Finally, we removed the guidance signal from the fusion module during the registration training process. As can be seen in the fourth column of Table~\ref{ablation} all the four metrics have undergone only a slight decrease and the network is still able to perform most of the registration. This suggests that the guidance signals, given by the fusion module, are mainly local details of the registration and are able to refine the BusReF performance.
\section{Conclusion}

In this paper, we proposed a bus-like architecture for training a multi-modal image registration and fusion network. The method, for the first time, unifies the features of multiple registration modules. The commonly used registration training and evaluation metrics are improved so that the network focuses on learning the registration and fusion in reconstructible regions. The experiments show that BusReF registered multi-modal images are more in line with the apriori knowledge based on human vision and are able to achieve higher registration accuracy. 

However, the proposed multi-modal image registration still has a lot of room for development. Although BusReF achieves feature reuse by mounting multiple registration modules to a bus, the gradient during training cannot be carried through the whole network. Perhaps end-to-end multi-modality registration algorithms will emerge in the future, which will further simplify the process and improve accuracy. Furthermore, is it possible for the results of multi-modality registration and fusion to be solved completely by end-to-end training? Although it is difficult, it is a very meaningful and promising topic. If a fully end-to-end output of registration and fusion can be achieved, it will be possible to find the correlation between the features for registration and the features for fusion.